\documentclass[letterpaper, 10pt, times, preprint]{elsarticle}

\usepackage{hyperref}

\usepackage[version=4]{mhchem}

\usepackage{lipsum}

\usepackage{tgtermes}
\usepackage{tikz}
\usepackage[letterpaper, portrait, margin=0.60in]{geometry}
\usepackage{hhline}
\usepackage{float}
\usepackage{multirow}
\usepackage{graphicx,booktabs}
\usepackage{tabularx}
\usepackage{xfrac}
\usepackage[compact]{titlesec}
\usepackage{wrapfig,booktabs}
\usepackage{mwe}
\usepackage{adjustbox}
\usepackage{makecell}
\usepackage{rotating}
\usepackage{overpic}
\usepackage{pict2e}
\usepackage{amsmath,array, amssymb}
\usepackage{xcolor,soul}
\usepackage{tcolorbox}
\usepackage{calrsfs}
\usepackage{longtable}
\usepackage{ragged2e}
\usepackage{siunitx}
\usepackage{tikz}
\usepackage{gensymb}
\usepackage{svg}
\usepackage{graphicx} 
\usepackage[frozencache, cachedir=.]{minted}
\usepackage{xcolor}

\usepackage{float}

\hypersetup{colorlinks,breaklinks,
            urlcolor=Maroon,
            linkcolor=Maroon}

\titlespacing*{\subsection}{0pt}{*0}{*0}
\titlespacing*{\subsubsection}{0pt}{*0}{*0}

\begin{document}

\begin{frontmatter}

    \title{\large Supply Risk-Aware Alloy Discovery and Design}

    \author{Mrinalini Mulukutla$^{a}$}
    \corref{mycorrespondingauthor}
    \ead{mrinalini.mulukutla@tamu.edu }
     \author{Robert Robinson$^{a}$}
    \author{Danial Khatamsaz$^{a}$}
    \author{Brent Vela$^{a}$}
    \author{Nhu Vu$^{a}$}
    \author{Raymundo Arróyave$^{a,b}$}
    
    \address{$^a$Materials Science and Engineering Department, Texas A\&M University, College Station, TX, USA 77840}
    \address{$^b$Mechanical Engineering Department, Texas A\&M University, College Station, TX, USA 77840}





\begin{abstract}
Materials design is a critical driver of innovation, yet overlooking the technological, economic, and environmental risks inherent in materials and their supply chains can lead to unsustainable and risk-prone solutions. To address this, we present a novel risk-aware design approach that integrates \emph{Supply-Chain Aware Design Strategies} into the materials development process. This approach leverages existing language models and text analysis to develop a specialized model for predicting materials feedstock supply risk indices. To efficiently navigate the multi-objective, multi-constraint design space, we employ Batch Bayesian Optimization (BBO), enabling the identification of Pareto-optimal high entropy alloys (HEAs) that balance performance objectives with minimized supply risk. A case study using the MoNbTiVW system demonstrates the efficacy of our approach in four scenarios, highlighting the significant impact of incorporating supply risk into the design process. By optimizing for both performance and supply risk, we ensure that the developed alloys are not only high-performing but also sustainable and economically viable. This integrated approach represents a critical step towards a future where materials discovery and design seamlessly consider sustainability, supply chain dynamics, and comprehensive life cycle analysis.
\end{abstract}

\begin{keyword}
Supply Chain Risk Considerations, Alloy Design, Bayesian Methods, High throughput Methods
\end{keyword}

\end{frontmatter}






\section{Introduction}

In an age defined by rapid technological advancements and an ever-increasing global demand for innovation, the field of materials science stands at the forefront of transformative discoveries. The pursuit of novel materials, their innovative design, and their applications are central to technological progress and scientific advancement. From super-strong alloys to cutting-edge nanomaterials, the possibilities are limitless. The main advantage of new alloy development lies in the ability to create components with unique structural, compositional, and mechanical properties tailored to specific performance requirements. Extensive interest in various sectors, including biomedical, aerospace, energy, nuclear, and many others, led to a rise in the global use of metals to meet the growing demands of various high-performing applications \cite{Krausmann2017GlobalUse, doi/10.2760/386650}. 

The design and development of novel alloys should begin with a clear understanding of the end-use application requirements, including operational environment, desired properties, complexity of geometry to be manufactured, and scales of production. The alloy design process is a multi-stage process that integrates these application requirements with scientific and engineering principles to develop materials that meet these often-competing criteria. This process begins with the selection of compositional space based on its compatibility with end-use requirements and the addition of alloying elements to enhance specific properties. This is followed by the utilization of computational tools such as CALPHAD (CALculation of PHAse Diagrams) to predict phase stability, microstructural evolution, and other thermodynamic properties with the goal of property prediction before the synthesis of experimental alloys for initial testing. In other words, traditional alloy design methodologies often prioritize mechanical properties, thermal stability, corrosion resistance and other desired criteria to ensure optimal performance in application. 

However, recent global supply chain challenges and resultant disruptions have underscored the critical need to integrate supply risk considerations into the alloy design process \cite{Cann2021SustainabilityOpportunities,Graedel2022AlloyLists,Ibrahim2023ChallengesWaste}. This paradigm shift is essential to ensure both the economic viability and sustainable development of advanced materials. The interdisciplinary research of supply chains and sustainability has received extensive, yet gradual, attention \cite{Cann2021SustainabilityOpportunities, Alonso2007MaterialResponses, Bhuwalka2023QuantifyingChains}; compared to the rapid global economic growth, sustainable supply chain management has not been systematically explored yet. Recognition of the urgency for sustainable manufacturing and supply chains that support global sustainability has become increasingly important for organizations around the world. They have been extensively explored individually; however, limited efforts have been concentrated on systematically mitigating the risks involved in sustainable supply chain management \cite{Raabe2023TheAlloys, Lbre2019, Olivetti2022TheRecovery}. It has not only great theoretical significance but also positive practical significance in providing a framework for the operation of a sustainable service supply chain from a sustainable development point of view. 

\begin{figure}[H]
        \centering
    \includegraphics[width=1.0\textwidth]{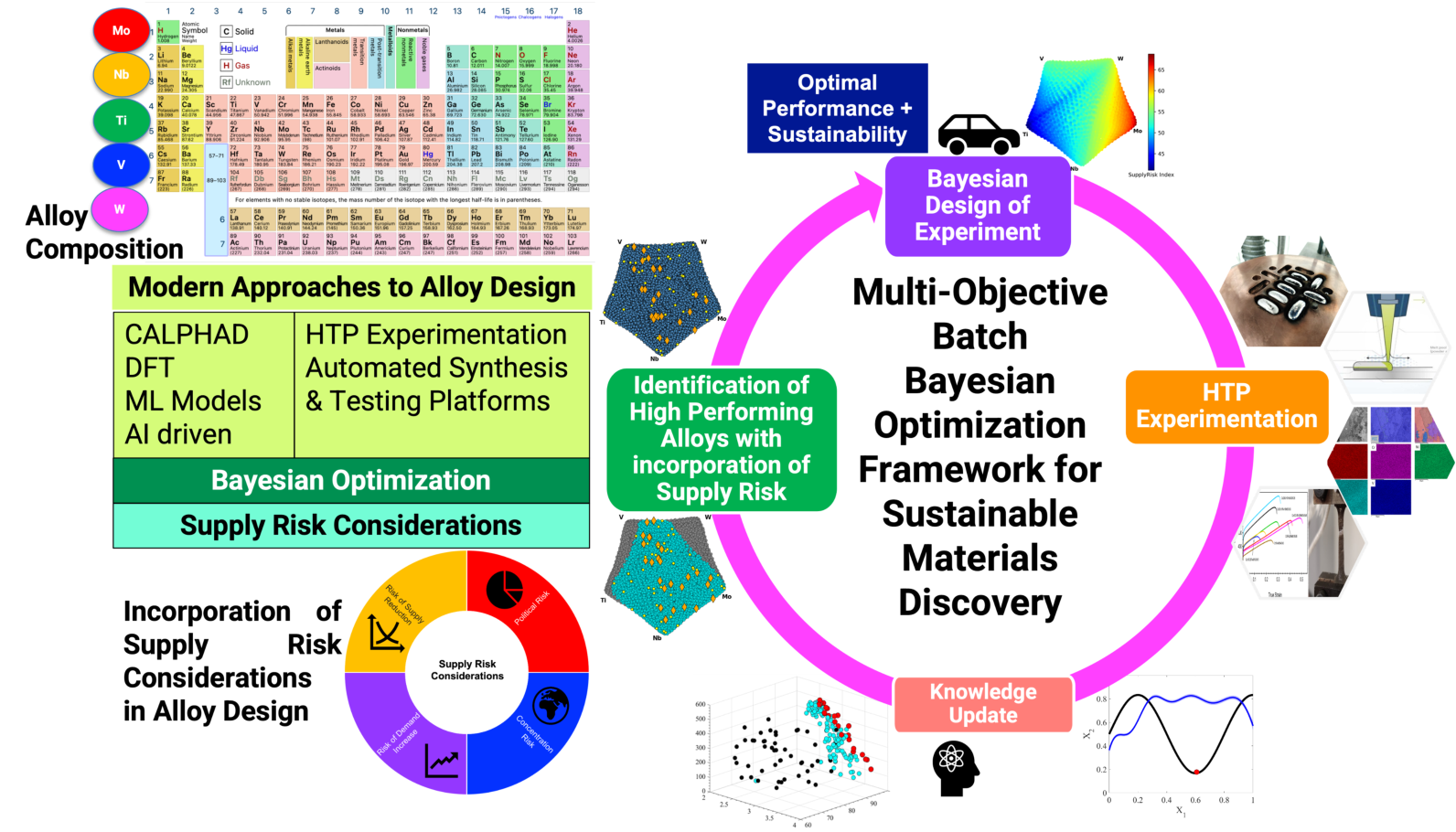}
        \caption{Overview of the main considerations in Alloy Design Workflow}
        \label{fig:Graphical_Workflow}
    \end{figure}

In the face of emerging challenges and ethical responsibilities, it has become abundantly clear that materials discovery and design cannot exist in isolation \cite{Schoenung2023SustainableEngagement}. Instead, these endeavors are profoundly intertwined with issues related to sustainability, supply chain dynamics, and full life cycle analysis \cite{Helbig2017BenefitsDevelopment, L.DouglasSmith2022AssessingConcepts}. To navigate this complex terrain, we must adopt a holistic approach where collaboration, innovation, and ethical consciousness converge. The comprehensive design and development process illustrated in \autoref{fig:Graphical_Workflow} visually represents the integration of end-use application requirements, computational methods, experimental validation, and supply risk considerations at each stage of alloy design. By systematically combining these elements, the alloy design process becomes more resilient and sustainable. This iterative approach, guided by a deep understanding of application requirements, advanced computational tools, and supply risk metrics, enables the development of innovative alloys that meet the stringent demands of modern engineering applications while ensuring material security and sustainability. By doing so, we can ensure that innovations not only transform the world but also secure a sustainable and ethical future for generations to come. 

Integration of sustainability and supply chain considerations in material design is not a novel concept, however, it has gained significant attention in the recent decades. Several researchers have made significant contributions to understanding how these factors can be systematically incorporated to bridge materials science with sustainability \cite{Helbig2016SupplyPhotovoltaics, Helbig2018SupplyMaterials, Helbig2020SupplySuperalloys, Rodby2023MaterialsBatteries, WangXinyiandKramer2022Sustainability-BasedAlloys}. Their works have focused on developing methodologies to evaluate the environmental and economic impacts of material choices across their life cycles by integrating principles from industrial ecology and materials science. This research emphasizes the importance of considering the full life cycle of materials, from extraction and processing to use and end-of-life disposal, ensuring that environmental impacts are addressed comprehensively rather than in isolation \cite{Schoenung2023SustainableEngagement}. Additionally, other works complement this research by developing tools and models that facilitate the assessment of material sustainability and supply chain risks \cite{Olivetti2022TheRecovery, Bhuwalka2023APredictions, Helbig2021AnAssessments, Adewusi2024THESTUDIES}. The concept of life cycle analysis (LCA) has significantly evolved since its inception. Initially, LCA studies focused primarily on energy consumption and greenhouse gas emissions related to material production. Today, LCA encompasses a wider range of environmental and social impacts, including resource depletion, toxicity, and social implications. The adoption of LCA in materials science is driven by the need to understand the cumulative impacts of material choices, which is crucial for developing strategies that minimize negative environmental and social outcomes \cite{Mancini2018CharacterizationEurope}. Integrating LCA into materials design has enabled researchers to identify more sustainable alternatives and optimize material systems for both performance and sustainability. Few other ongoing efforts further underscore the growing importance of incorporating sustainability and supply chain risks into the development of novel alloys, specially in the field of high entropy alloys (HEAs). A notable example is a study that proposed a rapid assessment framework to guide the search for sustainable alloys \cite{GORSSE2024e00938}. Their framework categorized sustainability indicators into three main areas including economic viability, environmental impact and human well-being. The study advocates for early consideration of societal impacts in alloy design, emphasizing that while trade-offs maybe necessary, efforts should always be made to minimize undesirable effects. Ultimately, their work attempted to provide a valuable tool for guiding sustainable alloy development, encouraging selection of elements that favor sustainability and reduce risks associated with supply chain. 

Supply chain indices for alloys can be complex in nature involving multiple risks at several stages in the process, including availability/cost of raw materials, complexity of manufacturing processes, environmental and geopolitical risks associated with the location resulting in instability, leaving the development process vulnerable to numerous threats \cite{Helbig2021AnAssessments}. The supply risk of critical elements, such as rare earth metals, has become a significant concern for industries reliant on stable and predictable material availability. These challenges necessitate the development of alloys that not only meet performance criteria but also mitigate supply chain vulnerabilities. To address these multifaceted challenges, this paper introduces a novel framework for supply risk-aware alloy design. The proposed approach leverages Bayesian Optimization (BO), an efficient optimization strategy known for its capability to handle complex and multi-objective problems with high uncertainty. By integrating supply risk metrics into the BBO framework, we aim to identify alloy compositions that balance desirable material properties with reduced dependency on high-risk elements.

Our framework starts with a thorough evaluation of supply risk for various alloying elements, considering factors such as global production distribution, recycling potential, geopolitical risks, and several others detailed in the following sections \cite{Helbig2020SupplySuperalloys}. A combined metric---the supply risk index---is then integrated into the BO algorithm, steering the search toward compositions that minimize supply risk while achieving the desired performance metrics. We validate the efficacy of our approach through a case study on a refractory multi-principal element alloy system, emphasizing the trade-offs and synergies between performance and supply risk. By pushing the boundaries of alloy design with an emphasis on supply risk, this research seeks to contribute to the development of resilient and sustainable material systems. Integrating supply risk considerations into the early stages of alloy design not only enhances material security but also aligns with broader objectives of sustainable resource management and environmental stewardship.

\section{Methods}
\label{Methods}

Alloy design has undergone remarkable evolution over the past few decades, propelled by advancements in computational techniques, experimental methodologies, and a more profound understanding of materials science. The creation of novel alloys with tailored properties is essential across a wide spectrum of industries, including aerospace, automotive, electronics, and biomedical devices. Contemporary alloy design leverages state-of-the-art technologies and interdisciplinary strategies to engineer materials that fulfill precise performance requirements. This article explores some of the key modern approaches to alloy design.

\subsection{Supply Risk Assessment Method}

The evaluation method employed to assess supply chain risk in this study follows the approach by Helbig et al. \cite{Helbig2020SupplySuperalloys}, as detailed in their articles \cite{Helbig2016SupplyPhotovoltaics,Helbig2018SupplyMaterials}. Their methodology uses a supply chain risk measurement framework categorized into four groups with twelve indicators at the elemental level to identify risks with equal weighting. 

Firstly, risks associated with supply reduction are quantified using the static range of reserves (S1), static range of resources (S2), and the secondary production share from old scrap (S3). Anticipated future technology demand (D1), the proportion of coupled production (D2), sector competition (D3), and substitutability (D4) contribute to risks driven by increasing demand. Market concentration risks arise from the concentration of producing countries (C1) and producing companies (C2). Political risks are assessed based on the stability of producing countries (P1), their policy perception (P2), and regulatory risk (P3).

\begin{figure}[H] 
        \centering
        \includegraphics[width=0.7\textwidth]{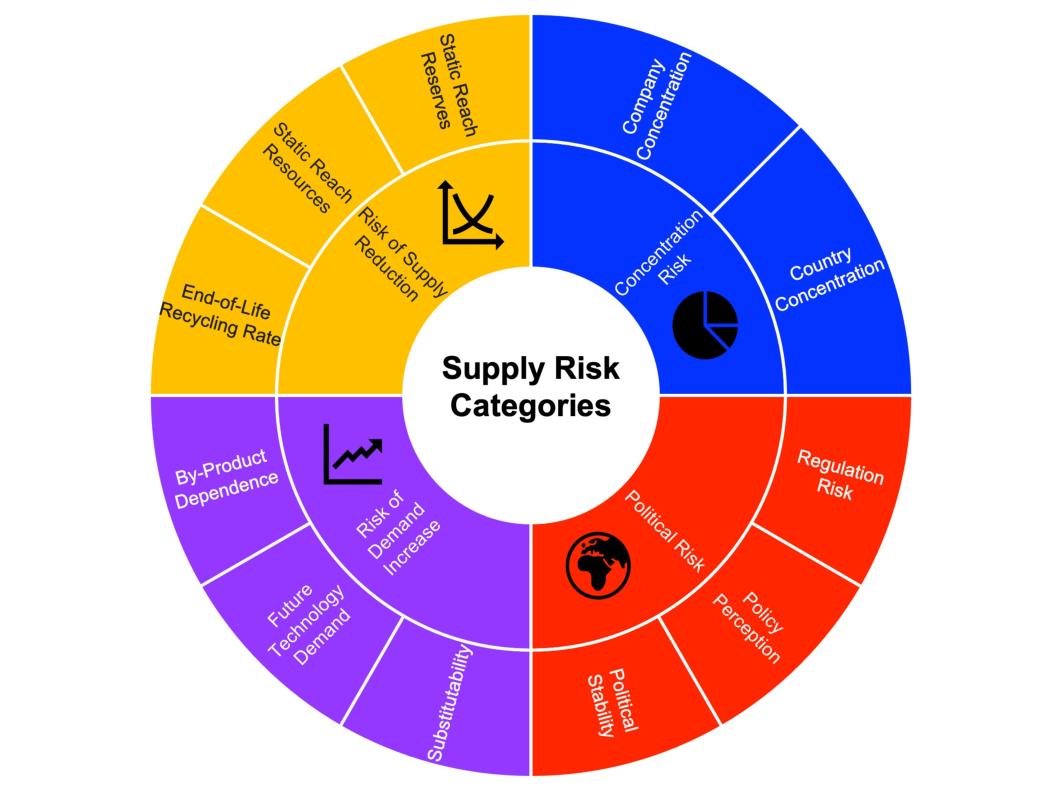}
        \caption{Supply Risk Categories}
        \label{fig:Supply_Risk_Categories}
    \end{figure}

This methodology is designed to highlight supply risks using relevant indicators for various superalloys. \autoref{fig:Supply_Risk_Categories} illustrates the different supply risk categories.

Additional details on each indicator, including calculations for various metals and normalization to a common scale of 0 to 100, are provided in the supplementary material \cite{Helbig2016SupplyPhotovoltaics,Helbig2018SupplyMaterials,Helbig2020SupplySuperalloys,Helbig2021AnAssessments}. The sum of the indicators at the alloy level, which aggregates the risk scores at the elemental level, yields a composite index that assesses the overall sustainability of the supply chain under consideration. 

Overall assessment of supply risk (SR) is given by the weighted mean of respective weights of indicators for each element given by the following expression:
       \begin{equation}
        \label{eq:kern}
         \sum SR_{mean} = \sum_{i\epsilon Alloy}SR_{i}/\sum_{i\epsilon Alloy}1
        \end{equation}
         
This supply risk assessment methodology evaluates alloys in relation to their key performance metrics based on the intended application. Alloys are considered more competitive if they can meet performance criteria without increasing supply risks, or if they offer comparable performance with significantly lower supply risks, assuming similar densities, environmental properties, and costs. However, when superior performance in a given application results in higher supply risks, it introduces a trade-off that necessitates further analysis.

\subsection{Modern Approaches to Alloy Design}

With the advent of computational tools and high throughput experimentation, modern approaches leverage them to accelerate the discovery and optimization of novel materials. Integrating supply risk consideration metrics into alloy design is a natural progression as a next step. 

\subsubsection{High Throughput Computational Tools and Experimentation}

Traditional methods of alloy development, which predominantly relied on empirical testing and trial-and-error experimentation, have increasingly been supplanted by advanced, data-driven, and computational approaches. These innovative frameworks not only expedite the alloy development process but also facilitate the design of alloys with tailored properties to meet specific application demands. Computational alloy design harnesses robust computational tools to predict the properties and behaviors of alloys prior to their experimental synthesis, thereby substantially reducing the time and costs associated with conventional trial-and-error methodologies.
A significant advancement in alloy design is the adoption of computational methods. Techniques such as Density Functional Theory (DFT) and CALPHAD (CALculation of PHAse Diagrams) allow for the prediction of properties and phase stability of potential alloy compositions prior to synthesis \cite{Rettenmayr2009AlloyTools, ArroyaveRaymundoandLi2016AlloyApproaches}. These methods, by modeling atomic-level interactions, provide insights into the microstructural evolution and mechanical behavior of alloys, thereby reducing the need for extensive experimental testing and conserving both time and resources.

Another key development in modern alloy design is the integration of machine learning and artificial intelligence (AI) \cite{Kalidindi2015MaterialsOutlook, Liu2017MaterialsLearning}. Machine learning algorithms, trained on extensive datasets of material properties, can reveal patterns and correlations that might be overlooked through traditional methods \cite{Huang2019Machine-learningAlloys, Wen2019MachineProperty, Catal2024MachineProperties}. These algorithms can suggest new alloy compositions with a high likelihood of achieving desired properties, thereby streamlining experimental efforts. The use of AI in alloy design has opened new avenues for rapidly discovering high-performance materials with unprecedented efficiency.

In addition to computational and AI-driven methods, experimental validation remains crucial in alloy design. While computational predictions and machine learning models can identify promising compositions, actual synthesis and testing are essential for verification. High-throughput experimentation, involving the rapid synthesis and evaluation of numerous alloy compositions, allows efficient exploration of vast compositional spaces. Automated systems with robotic platforms and advanced characterization tools significantly accelerate material property assessments \cite{Vecchio2021High-throughputHT-READ, Sokol2022AutomatedProduction}. These systems can conduct multiple experiments simultaneously, expediting alloy discovery. By integrating high-throughput methods with computational predictions, researchers can swiftly identify candidates for further investigation \cite{KhurramShahzad2024AcceleratingAdvances, Pyzer-Knapp2022AcceleratingRobotics}.

Integrated Computational Materials Engineering (ICME) represents a systematic approach to alloy design by integrating computational modeling, experimental validation, and industrial application \cite{Allison2011IntegratedSteps, Gautham2024AcceleratedPlatforms}. ICME frameworks bridge multiple length scales, from the atomic to the macroscopic, to predict material behavior under specific conditions. By combining detailed models of material processing, microstructure, properties, and performance, ICME enables the precise design of alloys optimized for targeted applications. This approach significantly reduces both the time and cost of alloy development while ensuring that the engineered materials possess well-defined, application-specific properties. The iterative process of prediction, synthesis, and rigorous validation within ICME guarantees that the final alloy compositions meet exacting performance criteria.

\subsubsection{Visualisation of High Dimensional Spaces}
Visualizing the high-dimensional spaces is inherently challenging due to the complexity and the vast amount of data involved. Uniform Manifold Approximation and Projection (UMAP) is a dimensionality reduction algorithm (DRA) that has been developed recently to address common challenges in data visualization. UMAP visualizations help in understanding the distribution and clustering of chemistry-structure, chemistry-property, and chemistry-performance relationships within the dataset, providing clear insights into the variability and central tendencies among different alloy compositions. The methodology for creating various high dimensional visualizations by Vela et al \cite{VelaVisualizingPractices, Vela2024VisualizingPractices} has been used extensively in our study. 

\subsubsection{Bayesian Optimization in Alloy Design}

The design of novel alloys with targeted properties is a complex, multidimensional challenge, often requiring the exploration of vast compositional spaces. Traditional trial-and-error methods, though historically important, are inefficient and resource-intensive. Bayesian Optimization (BO) has emerged as a powerful strategy to accelerate the discovery and optimization of new alloys by systematically guiding the search process. BO is a sequential design approach ideal for the global optimization of black-box functions, particularly when these functions are costly to evaluate. This method is particularly beneficial when the objective function is unknown, noisy, or expensive to assess—conditions often encountered in materials design. In alloy development, BO effectively navigates high-dimensional compositional spaces, identifying compositions that achieve desired properties \cite{Arroyave2022ADesign}.

Key components of BO include the surrogate model, the acquisition function—which dictates the selection of the next experiment—and the sequential evaluation process. The surrogate model, often a Gaussian Process (GP), provides a probabilistic representation of the objective function, offering predictions with uncertainty estimates at unobserved locations~\cite{Rasmussen2005GaussianLearning}.

In the presence of $N$ previously evaluated design points: $\{\mathbf{X}_{N}, \mathbf{y}_{N}\}$, where $\mathbf{X}_{N} = (\mathbf{x}_{1}, \ldots, \mathbf{x}_{N})$ are $N$ design points and $\mathbf{y}_{N} = \left(f(\mathbf{x}_{1}), \ldots, f(\mathbf{x}_{N})\right)$ are the respective outcomes, a GP representation at an unobserved location $\mathbf{x}$ is
\begin{equation}
\label{GP11}
f_{\textrm{GP}}(\mathbf{x}) \mid \mathbf{X}_{N}, \mathbf{y}_{N} \sim \mathcal{N}\left(\mu(\mathbf{x}),\sigma_{\textrm{GP}}^2(\mathbf{x})\right)
\end{equation}
where
\begin{equation}
\begin{aligned}
\label{meancov}
\mu(\mathbf{x}) &= K(\mathbf{X}_{N},\mathbf{x})^T[K(\mathbf{X}_{N},\mathbf{X}_{N})+\sigma^2_{n}I]^{-1} \mathbf{y}_{N}\\
\sigma_{\textrm{GP}}^2 (\mathbf{x})  &=  k(\mathbf{x},\mathbf{x}) - K(\mathbf{X}_{N},\mathbf{x})^T\\
&\qquad \quad [K(\mathbf{X}_{N},\mathbf{X}_{N})+\sigma^2_{n}I]^{-1}K(\mathbf{X}_{N},\mathbf{x})
\end{aligned}
\end{equation}
where $k$ is a real-valued kernel function defined over the input space to account for correlation between design points, $K(\mathbf{X}_{N},\mathbf{X}_{N})$ is a $N \times N$ matrix with $m,n$ entry as $k(\mathbf{x}_{m}, \mathbf{x}_{n})$, and $K(\mathbf{X}_{N}, \mathbf{x})$ is the $N \times 1$ vector with $m^{th}$ entry as $k(\mathbf{x}_{m}, \mathbf{x})$ that defines the correlation of the unobserved point with the $m^{th}$ observed design point. The term $\sigma^2_{n}$ models the experimental error associated with each observed design point. 

The squared exponential covariance function is a common choice for constructing a GP:

\begin{equation}
\label{eq:5}
k(\mathbf{x},\mathbf{x'}) = \sigma_s^2 \exp\left(- \sum_{h = 1}^{d} \frac {(x_h-x'_h)^2}{2l_h^2}\right )
\end{equation}

where $d$ represents the input dimensions, $\sigma_s^2$ is the GP uncertainty (signal variance), and $l_h$, with $h = 1,2,\ldots,d$, denotes the characteristic length-scale, determining correlation strength in each dimension. Although the squared exponential kernel is widely used for its smoothness and flexibility, kernel choice depends on the specific problem. Kernel engineering is an active field, with significant research focused on developing kernels that better capture system physics. Recently, there has been growing interest in physics-informed kernel selection, where domain knowledge is integrated into kernel design to enhance model accuracy and interpretability \cite{khatamsaz2023physics,vela2023data}.

The acquisition function \cite{frazier2018tutorial} is crucial in Bayesian optimization, balancing exploration of new regions with exploitation of current knowledge. Common acquisition functions include Expected Improvement (EI), Probability of Improvement (PI), and Upper Confidence Bound (UCB). At each iteration, the acquisition function is maximized to select the most promising candidate for the next experiment. This sequential process updates the surrogate model with new observations, refining the search space. The workflow typically involves defining the objective function, initial sampling, surrogate model construction, acquisition function optimization, and iterative evaluation. This systematic approach minimizes the number of required experiments or simulations, conserving time and resources while accelerating the discovery of optimal material compositions.

Batch Bayesian Optimization (Batch BO) extends sequential BO by allowing multiple candidate evaluations per iteration \cite{couperthwaite2020materials,khatamsaz2021efficiently}. This approach is particularly advantageous in high-dimensional design spaces and complex objective functions, such as those in alloy design. Batch BO reduces optimization time through concurrent evaluations, crucial in resource-intensive materials science experiments. Additionally, by sampling diverse points within each batch, Batch BO enhances exploration, reduces the risk of convergence to local optima, and improves robustness by mitigating the impact of noisy observations. These features make Batch BO an effective strategy for accelerating the discovery of optimal material compositions.

Bayesian Optimization frameworks have been successfully applied in various alloy design scenarios \cite{Pedersen2021BayesianReduction, Sulley2024AcceleratingLearning, Mamun2024AcceleratedOptimisation, Enyoghasi2021BayesianAssessment, Bhuwalka2023APredictions, Khatamsaz2022Multi-objectiveAlloys}. For instance, researchers have used Bayesian design to iteratively refine the chemical composition and processing of high-strength steel, achieving significant improvements in tensile strength and elongation \cite{Gong2024IterativeOptimization}. In high-entropy alloys (HEAs), BO has efficiently navigated the multi-component compositional space to discover new compositions with exceptional mechanical properties \cite{Sulley2024AcceleratingLearning}. Khatamsaz et al. \cite{Khatamsaz2023BayesianApproach} extended these methods to solve a multi-objective design problem in the Mo-Nb-Ti-V-W system, a representative Multi-Principal Element Alloy (MPEA) for next-generation turbine blades, showing greater efficiency in identifying optimal compositions for lightweight alloys compared to brute-force methods. Despite these successes, challenges remain, particularly in handling high-dimensional spaces, necessitating the development of scalable BO algorithms and the integration of multi-fidelity models to enhance efficiency and accuracy. Incorporating domain-specific knowledge and constraints into BO frameworks and advancing real-time BO that adapts dynamically to experimental data are promising directions for accelerating alloy discovery. Overall, BO offers significant advantages in efficiency, adaptability, and flexibility, systematically guiding the rapid development of novel alloys with tailored properties to meet modern engineering demands\cite{Arroyave2022ADesign}.
    
\subsubsection{Indicator Extraction using Large Language Models}

The acquisition of data required to compute the supply risk of a given element is a time-consuming process. The statistics that inform supply risk indicators are derived from documents, periodicals, and research papers, which vary in length, language, and format. Traditionally, accurate data collection necessitates manual searching and sorting by a human to extract relevant information and compile it into a usable table for research purposes. However, recent advancements in Large Language Models (LLMs) and Natural Language Processing (NLP) have enabled the automation of this task. A workflow leveraging LLMs has been defined, and an application is currently under development. The objective of this workflow is to minimize human involvement in data collection for supply risk indicators while ensuring accuracy and allowing for easy modifications to the list of indicators used in calculating the supply risk for a material. The method for data extraction is twofold: deterministic text search algorithms are employed to identify areas where relevant statistics may be found, followed by the use of a Large Language Model (LLM) for Natural Language Processing (NLP) to perform binary classification and determine the actual important data.    

    \begin{figure}[H] 
        \centering
        \includegraphics[width=1.0\textwidth]{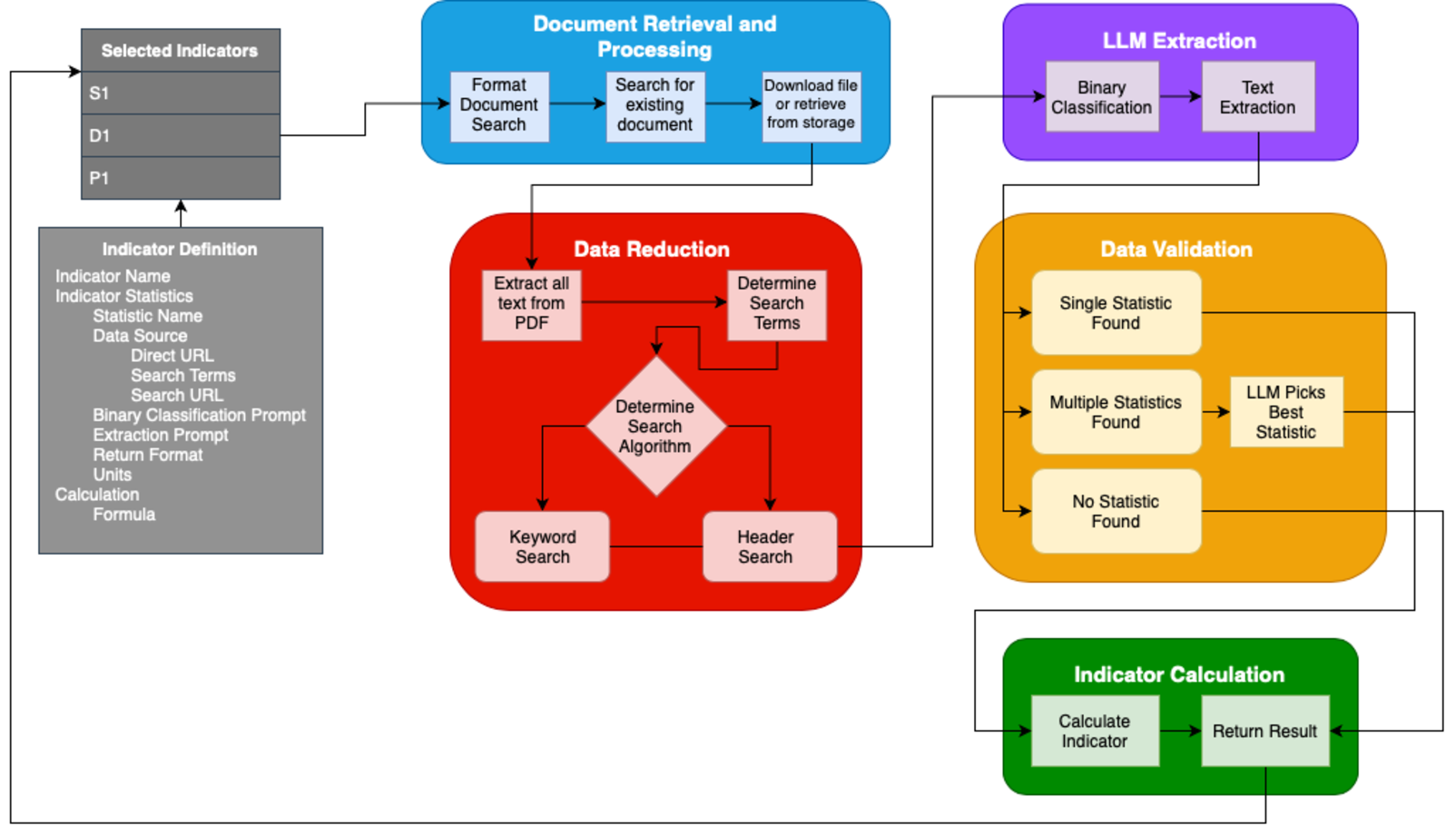}
        \caption{A high-level overview of the main steps of the text extraction workflow. Indicators are defined, which are then used for data collection. The collected data is filtered and reduced before going to a LLM for text extraction and data validation.}
        \label{fig:LLM workflow}
    \end{figure}

The workflow for data extraction consists of four primary steps: Document Retrieval and Processing, Data Reduction, Large Language Model (LLM) Processing, and Data Validation and Error Handling.

\emph{Document Retrieval and Processing} begins by retrieving relevant documents for a specific indicator. Documents can be accessed directly through known links or by searching with keywords and URLs. This step is particularly crucial when the data source is predefined, such as the United States Geological Survey's Material Commodity Survey. In this case, the document link is iteratively checked, starting from the current year, until the latest available version is found. This approach ensures that the most current data is extracted while avoiding redundant downloads. Ongoing development is focused on a search-based retrieval method that aims to identify relevant data from non-periodical sources, such as newly published research papers, by leveraging keyword searches and text analysis techniques.

\emph{Data Reduction} follows, where the retrieved document is stripped down to the relevant text to minimize the number of calls made to the LLM, a significant bottleneck in the workflow. This step begins with collecting keywords related to the indicator, which are then used to search for pertinent sections within the document. The workflow includes mechanisms to handle language differences by translating keywords if the document is in a different language. Two text search methods are employed: a keyword proximity search that identifies clusters of relevant terms and a header search that targets specific headers in documents where statistics are typically tabulated.

In the \emph{LLM Processing} stage, the identified text sections are passed to an LLM for binary classification. The LLM determines whether each text segment contains relevant statistics, filtering out irrelevant data. If a segment is confirmed as relevant, the LLM extracts the required data in a predefined format, ensuring accurate and consistent data extraction.

\emph{Data Validation and Error Handling} is the final critical step in the workflow. This stage addresses potential issues such as multiple statistics being found or no relevant data being extracted. Depending on the scenario, the workflow either filters out unnecessary statistics, prompts manual document review, or adjusts prompts for better accuracy. Three primary cases are managed: when a single statistic is correctly extracted, when multiple statistics need to be consolidated, and when no relevant statistics are identified. The validated data is then used for further calculations.

The workflow culminates in the \emph{Indicator Calculation} step, where the extracted statistics are used to compute indicator values. These calculations follow the formulas defined within each indicator's specifications, and the results, along with logs of the extraction process, are presented to the user.

Indicators, defined with fields for name, statistics, and calculations, are central to this workflow. They allow users to control the calculation of supply risks for materials with minimal manual intervention. Indicators are designed to be flexible and adaptable over time, supporting various materials and elements through dynamic text and URL adjustments. While constructing these indicators requires significant development and fine-tuning, the outcome is a largely autonomous system capable of extracting relevant statistics and calculating indicators across a wide range of scenarios. A comprehensive explanation of each step in this workflow, along with specific implementation details and code snippets, is provided as supplemental information.

\subsection{Incorporating Supply Chain Risk Considerations in Alloy Discovery and Development - A Case Study for MPEA}

Refractory Multi-Principal-Element Alloys (MPEAs) are increasingly recognized for their exceptional high-temperature stability and mechanical properties, making them prime candidates for advanced applications in aerospace, energy, and other high-performance sectors\cite{vela2023high,Khatamsaz2022Multi-objectiveAlloys}. These alloys, characterized by their complex compositions and robust performance under extreme conditions, are particularly well-suited for demonstrating innovative approaches in alloy discovery and development. 

In this study, we focus on incorporating supply chain risk considerations into the design process of MPEAs targeting MoNbTiV system, utilizing a multi-objective Batch Bayesian optimization (MOBBO) framework to balance performance and sustainability. We leverage modern approaches to alloy design to solve a hypothetical problem where materials performance \emph{and} supply risk concerns are taken into account and illustrate the effectiveness of incorporating supply risk considerations into the alloy design process using a multi-objective Batch Bayesian optimization (MOBBO) framework\cite{Khatamsaz2022Multi-objectiveAlloys,khatamsaz2021bayesian}. 

The motivation for this alloy system stems from the critical need to address supply chain risks associated with the elements used in alloy formulations. Refractory High Entropy Alloys (RHEAs), with their diverse and often rare elemental constituents, face significant challenges related to resource availability and market volatility. By integrating supply risk indices into our optimization framework, we aim to develop alloy designs that are not only high-performing but also resilient against supply disruptions and environmental concerns. To illustrate the supply risk aware design strategy, we have used the supply risk assessment methodology described to determine the overall supply risk index for refractory alloy space in search of Pareto-optimal constraint-satisfying alloys for turbine applications. 

\subsubsection{Definition of Design Problem for Case Study}
The design of structural materials for turbine application is a highly complex and constrained problem, requiring the simultaneous satisfaction of multiple objectives and constraints. Candidate Alloys must satisfy these constraints and objectives to ensure their manufacturability and functionality. Our study targets the MoNbTiVWb system with constraints and objectives summarized in \autoref{tab:constraint_objective}. In addition to these constraints, the alloy must exhibit high strength at elevated temperatures to carry the necessary structural loads during operation while possessing sufficient ductility at room temperature to minimize the risk of fracture. Additionally, the density of the alloys must be minimized to meet the lightweight requirements essential for aerospace applications, and thermal conductivity must be maximized to ensure efficient thermal management. 

     \begin{table}[h!]
        \centering
        \begin{tabular}{|l|l|}
        \hline
            Property & Constraint\\
            \hline
            Melting Point & $T_{m}$ $<$  3000$^{\circ}$C\\
            Phase Stability & BCC \\
            Coefficient of Thermal Expansion & $<$ $10^{-5} K^{-1}$\\
            \hline
            Property & Objective\\
            \hline
            HT Yield Strength & Max $\sigma _{HT}$\\
            Ductility & Max $C_{11} - C_{22}$\\
            Ductility & Max B/G\\
            Thermal Conductivity& Max $\kappa$\\
            Density & Min $\rho$\\
            Supply Risk Index & Min $SR_{mean}$\\
            \hline
        \end{tabular}
        \caption{Constraints and Objectives chosen for design problem in this study}
        \label{tab:constraint_objective}
    \end{table}
    
The material system Mo-Nb-Ti-V-W and all its potential subsystems from 2 to 5 elements were sampled with a resolution of 5 atomic \%, resulting in an elemental phase space comprising 10,621 compositions. To streamline for feasibility, constraints were designed to filter the compositional space in an intelligent manner. They include avoiding alloys with high melting points ($>$ 3000 $\circ$C) for manufacturability, maintaining a small linear thermal expansion coefficient ($<$ $10^{-5} K^{-1}$) to minimize thermal stress during temperature fluctuations, and ensuring full body-centered cubic (BCC) phase stability to achieve desired mechanical properties for application. 
    
\subsubsection{Models for constraints}
Models for two of the three constraints used high-fidelity CALculation of PHAse Diagram (CALPHAD) based simulations. Thermo-Calc API, \emph{TC-Python}, was utilized to allow for batch querying of the coefficient of thermal expansion across the entire compositional space. CALPHAD method using ThermoCalc has proved to be effective in achieving accurate single phase solid solution predictions. BCC Phase stability of the alloys was predicted using Thermocalc’s TCHEA7 database, to determine compositions that would have a Body-Centered Cubic (BCC) crystal structure at operating temperature of 1300 °C \cite{Khatamsaz2023BayesianApproach}. Melting point was estimated using rule of mixtures and is computed for the entire compositional space for the purpose of this study. The constrained system with desired material phase and properties reduced the number of compositions to 7991 which is approximately 75\% of its original size. 

\subsubsection{Models for Objectives}
In this work, we evaluate the high-temperature (1300°C) yield strength objective using a physics-based model developed by Curtin and Maresca \cite{Maresca2020Mechanistic1900K}, which we consider the true model for this study. Ductility is approximated using fundamental material properties such as the Pugh ratio and Cauchy pressure. In metals, the Pugh ratio, defined as the ratio of bulk modulus to shear modulus (B/G), reflects the material's ability to undergo plastic deformation (B) without fracture (G) \cite{Pugh1954XCII.Metals}. Pettifor \cite{Pettifor1992TheoreticalIntermetallics} proposed Cauchy pressure as an indicator of intrinsic ductility or brittleness, defined as the difference between the elastic constants C12 and C44. A positive Cauchy pressure suggests non-directional metallic bonds, leading to intrinsic ductility, while a negative Cauchy pressure indicates directional bonds, resulting in brittleness. 

In this study, the DFT-based Korringa–Kohn–Rostoker Green’s function (DFT-KKR-CPA) method is used as the true model for both ductility objectives \cite{Khatamsaz2023BayesianApproach}. The evaluation methods follow the approach detailed by Khatamsaz et al. \cite{Khatamsaz2023BayesianApproach}. Similar to the constraints, CALPHAD-based simulations using the high-entropy alloy database TCHEA7 and Thermo-Calc's equilibrium simulations were employed to calculate the density and thermal conductivity across the entire compositional space. Finally, the supply risk index was calculated by applying the assessment methodology described earlier, estimating the weighted mean of respective indicator weights for each element. This index serves as the indicator for supply risk criteria in this study.

From the sampled compositions, a few alloys were selected for the first batch of trials. Rather than making these selections arbitrarily, we employed an algorithm called K-Medoids Clustering for a more systematic approach. K-Medoids clustering groups data into several clusters by identifying the central data point within each cluster using a nearest neighbor algorithm \cite{Park2009AClustering}. This space-filling method\cite{couperthwaite2020materials,vazquez2023deep} enables efficient exploration across various combinations of subsystems from as wide a range of systems as possible, ensuring representation across the diverse array of alloy systems. Using this technique, 10 starting candidate alloys were selected for the first iteration in each case study. These alloys were then used in the MOBBO framework to achieve multiple objectives: optimizing mechanical performance by balancing strength and ductility, maximizing thermal conductivity, minimizing density, and addressing sustainability by minimizing the supply risk index.

For comparative analysis within the framework of batch Bayesian optimization, this study is segmented into four distinct cases:
\begin{itemize}
    \item Perform MOBBO over the entire design space, targeting performance metrics exclusively.
    \item Perform MOBBO over the entire design space, incorporating both performance metrics and supply risk criteria.
    \item Perform MOBBO on the constrained feasible space, focusing solely on performance metrics.
    \item Perform MOBBO on the constrained feasible space, integrating both performance metrics and supply risk criteria.
\end{itemize}

The results of this study will include evaluating the system with and without supply risk criteria, assessing trade-offs in performance, calculating cost savings, and analyzing convergence efficiency. These results will demonstrate the significance of incorporating supply chain risk considerations into alloy design, highlighting how this approach can lead to more resilient and economically viable materials. 

\begin{figure}[H] 
        \centering
        \includegraphics[width=1.0\textwidth]{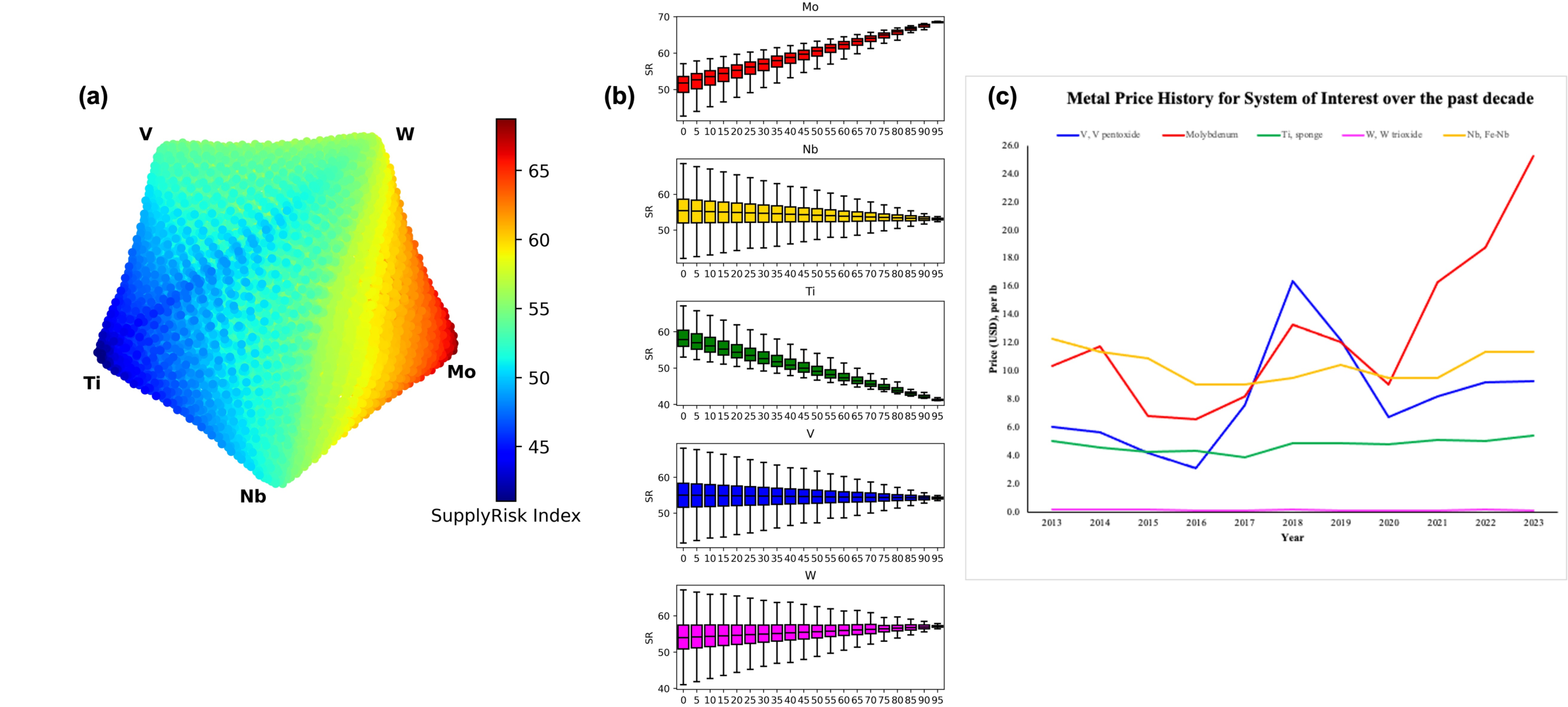}
        \caption{(a) UMAP visualisation for supply risk mean, (b) Box-whisker plots summarizing supply risk index as a function of individual alloying element concentrations, and (c) trends for price history of system of interest over the past decade}
        \label{fig:Supply_Risk_Visualisation}
    \end{figure}

For the system of interest, UMAP visualization \cite{Vela2024VisualizingPractices} and box-whisker plots of the supply risk index, based on simple arithmetic mean, indicate that Molybdenum has the highest SR mean, while Titanium has the lowest SR mean (\autoref{fig:Supply_Risk_Visualisation}(a,b)). This calculation is partially based on data from 2015 and is expected to increase due to changes in current socioeconomic conditions worldwide. Additionally, refractory metals carry an elevated risk due to their higher costs (\autoref{fig:Supply_Risk_Visualisation}(c)).

We then process this data and implement Batch Bayesian optimization with Gaussian process regression to model uncertainties and guide the search for optimal solutions. Information from various sources is integrated into the MOBBO\cite{Khatamsaz2022Multi-objectiveAlloys} framework for each case study, providing suggestions for the next batch of 10 candidates with the objective of improving the target performance metrics (objectives). By incorporating Bayesian insights and supply risk considerations, this study offers a robust framework for designing alloys that address both performance and sustainability challenges. Our work aims to advance sustainable materials science and provide practical solutions for mitigating supply chain risks in high-performance alloy development.

\subsubsection{Multi-Objective Batch Bayesian Optimisation Framework}

Multi-objective batch Bayesian optimization (MOBBO) has been chosen for the case study as it combines the principles of Bayesian optimization of multiple performance metrics with batch processing\cite{couperthwaite2020materials,KHATAMSAZ2022118133,Khatamsaz2022Multi-objectiveAlloys} to efficiently explore the design space and identify optimal solutions. 
Bayesian optimization is an iterative method used for optimizing expensive-to-evaluate, usually black-box type of functions. 

A multi-objective optimization problem is defined as
\begin{equation}
\label{eq:problem}
    \mbox{minimize}\,\,\,  \{f_1(\textbf{x}), ...,f_n(\textbf{x})\}, \textbf{x} \in \mathcal{X}
\end{equation}
where $f_1(\textbf{x}),\ldots, f_n(\textbf{x})$ are the objectives and $\mathcal{X}$ is the feasible design space. In multi-objective optimization, there is usually no single solution to optimize all objectives simultaneously. Rather, the optimal solution is a set of designs with trade-offs that form the Pareto front in the objective space. The optimal solutions of a $n$ objective design problem are represented as $\mathbf{y}\prec \mathbf{y'}$ and can be defined mathematically as
\begin{equation}
\begin{aligned}
   \{\mathbf{y}: \mathbf{y} & = (y_1, y_2, \ldots, y_n), \; y_i \leq y_i' \;\; \forall \; i \in \{1, 2, \ldots, n\}, &  \exists \; j \in \{1, 2, \ldots, n\} : y_j < y'_j\}
\end{aligned}{}
\end{equation} 
where $\mathbf{y}' = (y_1', y_2', \ldots, y_n')$ are possible objective outputs.  
The set of $\mathbf{y} \in \mathcal{Y}$, where $\mathcal{Y}$ is the objective space, is known as the Pareto front.

There are various techniques for estimating the Pareto front in multi-objective optimization problems, such as the weighted sum approach~\cite{Marler2010TheInsights}, the adaptive weighted sum approach~\cite{Kim2005AdaptiveGeneration}, normal boundary intersection methods~\cite{Das1998Normal-BoundaryProblems} and hypervolume indicator methods~\cite{Beume2009S-MetricProblem,Bradstreet2010ACalculations,Emmerich2011Hypervolume-basedComputation,Fonseca2006AnIndicator,Russo2012QuickHypervolume,Yang2007NovelSet,Zitzler1999MultiobjectiveApproach}, among others.

In multi-objective optimization, we aim to optimize multiple conflicting objectives simultaneously. The goal is to discover the Pareto front, which represents the set of non-dominated solutions. Non-dominance indicates that not all performance metrics (objectives) associated with a design are outperformed by any other design points. In MOBBO, each objective is modeled using a separate Gaussian process. For \textit{M} objectives, we have $f_{i}(x) \sim \textrm{GP}(\mu _{i}(\textbf{x}),k_{i}(\textbf{x},{\textbf{x}}')) \, \textrm{for} \, i=1,....,M$. Hypervolume improvement (HVI) is one of the key concepts in multi-objective optimization and provides a measure of the quality of solutions with respect to the optimal Pareto front. Hypervolume is defined as the volume of the objective space that is dominated by the Pareto front and bounded by a reference point. HVI quantifies the increase in hypervolume by adding a new non-dominated solution and is helpful in identifying the potential contribution of a yet-to-be-observed design toward improving the Pareto front\cite{zhao2018fast}. For a new candidate \textbf{x}, HVI is given by $\textrm{HVI(\textbf{x}) =} \textrm{HV}(P\cup {\textbf{x}}) - \textrm{HV}{(P)}$
where \textit{P} is the current Pareto front. 

In the sparse data regime, it is very risky to rely on a few observations on the system of interest to fit the predictive Gaussian Process (GP) used in BO as the GP hyperparameters would be highly dependent on the acquired samples. In this case, it is more advantageous to assume that, in the sparse data regime, all potential values of hyperparameters (within reasonable constraints) are feasible, \emph{a priori}. This assumption naturally leads to a batch modification to conventional BO algorithms, making the optimization process more robust by taking advantage of parallel execution capabilities\cite{couperthwaite2020materials}.

In this work, batch Bayesian optimization (BBO) extends the single-point acquisition to selecting a batch of points to evaluate in parallel. The batch acquisition function evaluates the utility of a set of points simultaneously. The function for a batch ${\textbf{x}_{1},\textbf{x}_{2},...,\textbf{x}_{n}} \, \textrm{is} \, \alpha ({\textbf{x}_{1},\textbf{x}_{2},...,\textbf{x}_{n}}) = \sum_{i=1}^{n}\textrm{HVI}(\textbf{x}_{i})$, guiding the selection of points that together improve the hypervolume most effectively.
Latin hypercube sampling (LHS) is a statistical method for generating a distribution of sample points in a multi-dimensional space. In MOBBO, LHS is used to generate sample points that spread uniformly across the design space. LHS divides each dimension into predetermined equally probable intervals and samples one point from each interval, ensuring good coverage of the design space. For each sample point \textbf{x}, we use Gaussian processes to predict the objective values and calculate the expected HVI. The next batch of candidates is selected by maximizing the batch acquisition function.
Gaussian processes are crucial in predicting the HVI associated with unobserved data points. For each candidate point \textit{x}, the Gaussian process provides a predictive distribution $\mathit{N(\mu (\textbf{x}),\sigma ^{2}(\textbf{x}))}$ for which the predictive mean and variance are given by the covariance vector/matrix for the \textit{i}-th objective. The hypervolume contribution of a point \textbf{x} is determined by integrating the predicted improvement over the Gaussian process distribution.

\begin{figure}[H] 
        \centering
        \includegraphics[width=1.0\textwidth]{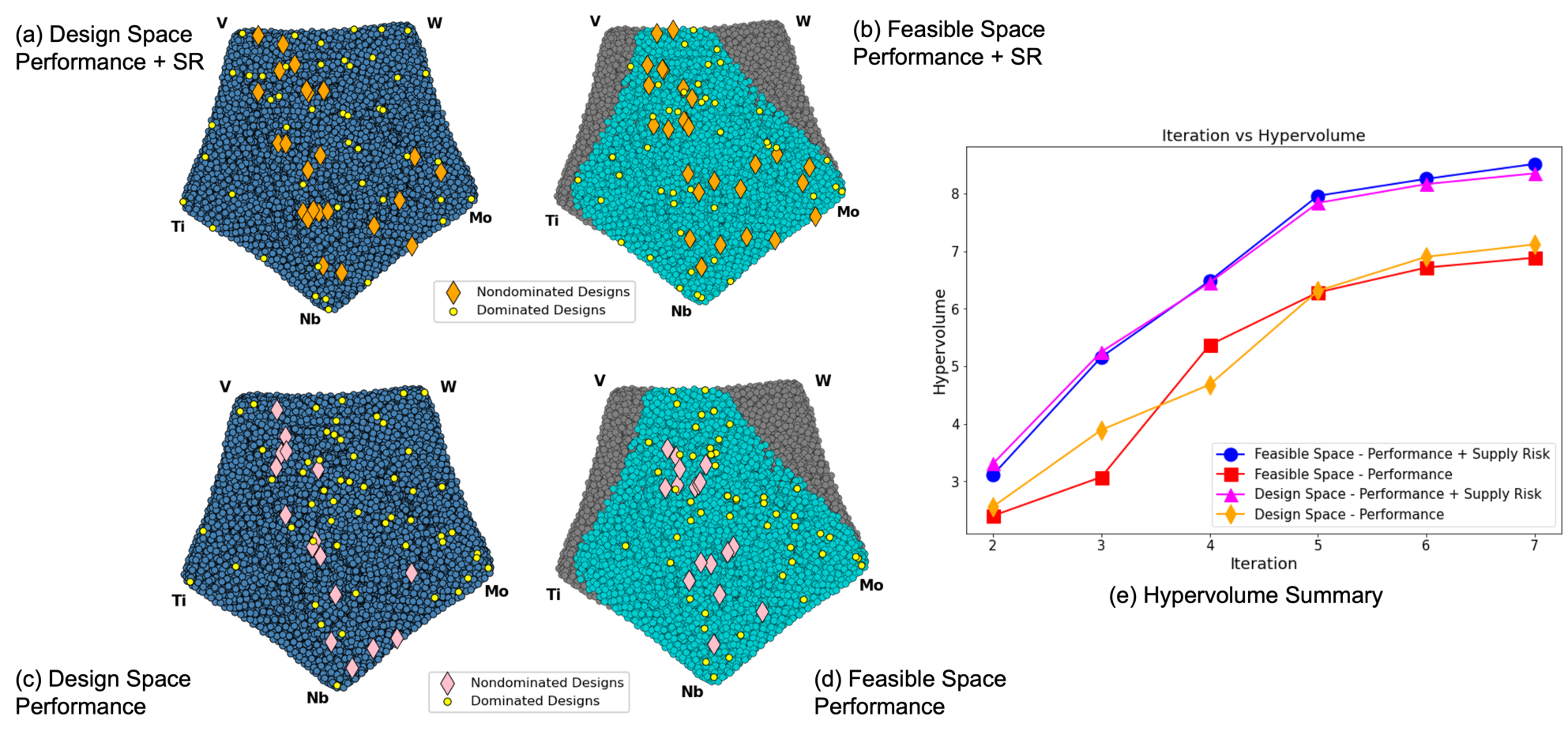}
        \caption{(a-d) Visualisation of Pareto Optimal Alloys for each case study with the BBO framework. Dominated designs are shown in yellow circles in each case. (e) Average Hypervolumes of data after each iteration showing minimal improvement after 5th iteration.}
        \label{fig:HV_Pareto}
    \end{figure}

The MOBBO algorithm starts with generating an initial set of points using LHS and evaluating their objective values. GPs are then trained for each objective using the evaluated points. In each iteration, candidate points are generated using LHS, and the batch acquisition function is evaluated for these points. The batch of points that maximizes the acquisition function is selected and evaluated, and the Gaussian processes are updated accordingly. This process is repeated until a convergence criterion is met, which is minimal improvement in the hypervolume for our study. The figure (\autoref{fig:HV_Pareto}) illustrates the concept of Expected Hypervolume Improvement (EHVI) in a two-objective minimization problem. The plot showcases existing Pareto-optimal solutions (grey points), a reference point (black), and three new candidate solutions with different scenarios: a dominated point (magenta) but with large uncertainty, a point that is dominated regarding objective 1 but expected to improve HV by improving objective 2 (blue), and a point with large confidence in generating a non-dominated point regardless of its uncertainty (green). 
The shaded areas represent the average hypervolume improvement for each candidate point according to GP predictions. Exact calculations of EHVI are needed to determine the potentials since the uncertainty plays a key role here.
Uncertainty in the objective values is visualized by Gaussian ellipses surrounding each candidate point.

\begin{figure}[H] 
        \centering
        \includegraphics[width=0.5\textwidth]{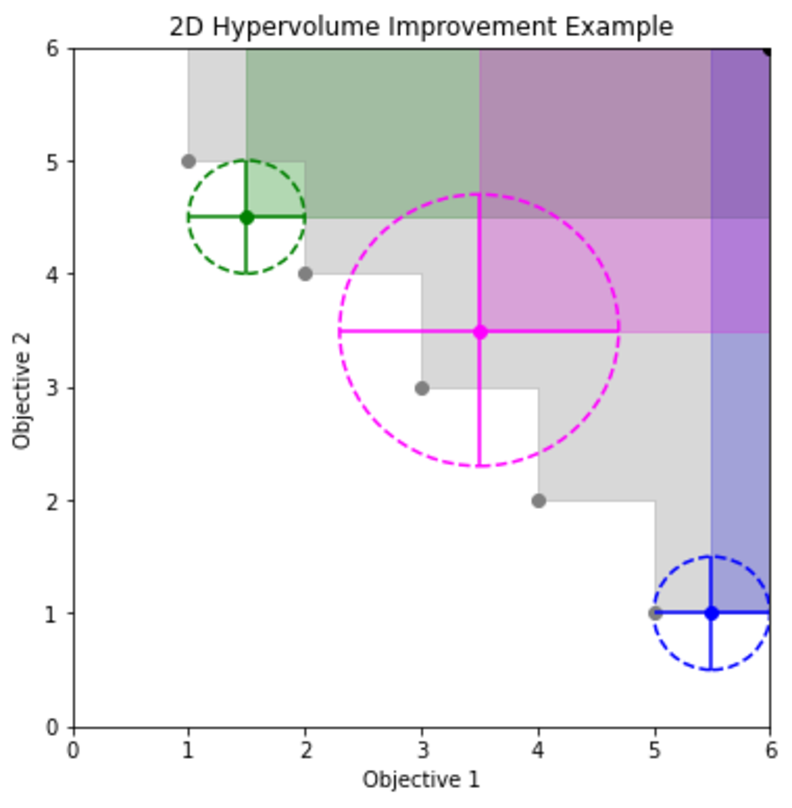}
        \caption{Schematic of a bi-objective optimization illustrating three candidate points, each with associated uncertainty ellipses, demonstrating varying levels of hypervolume improvement (HVI) relative to a reference point, with areas of improvement highlighted for each point.}
        \label{fig:BO}
    \end{figure}
\section{Results}

\begin{figure}[H] 
        \centering
        \includegraphics[width=1.0\textwidth]{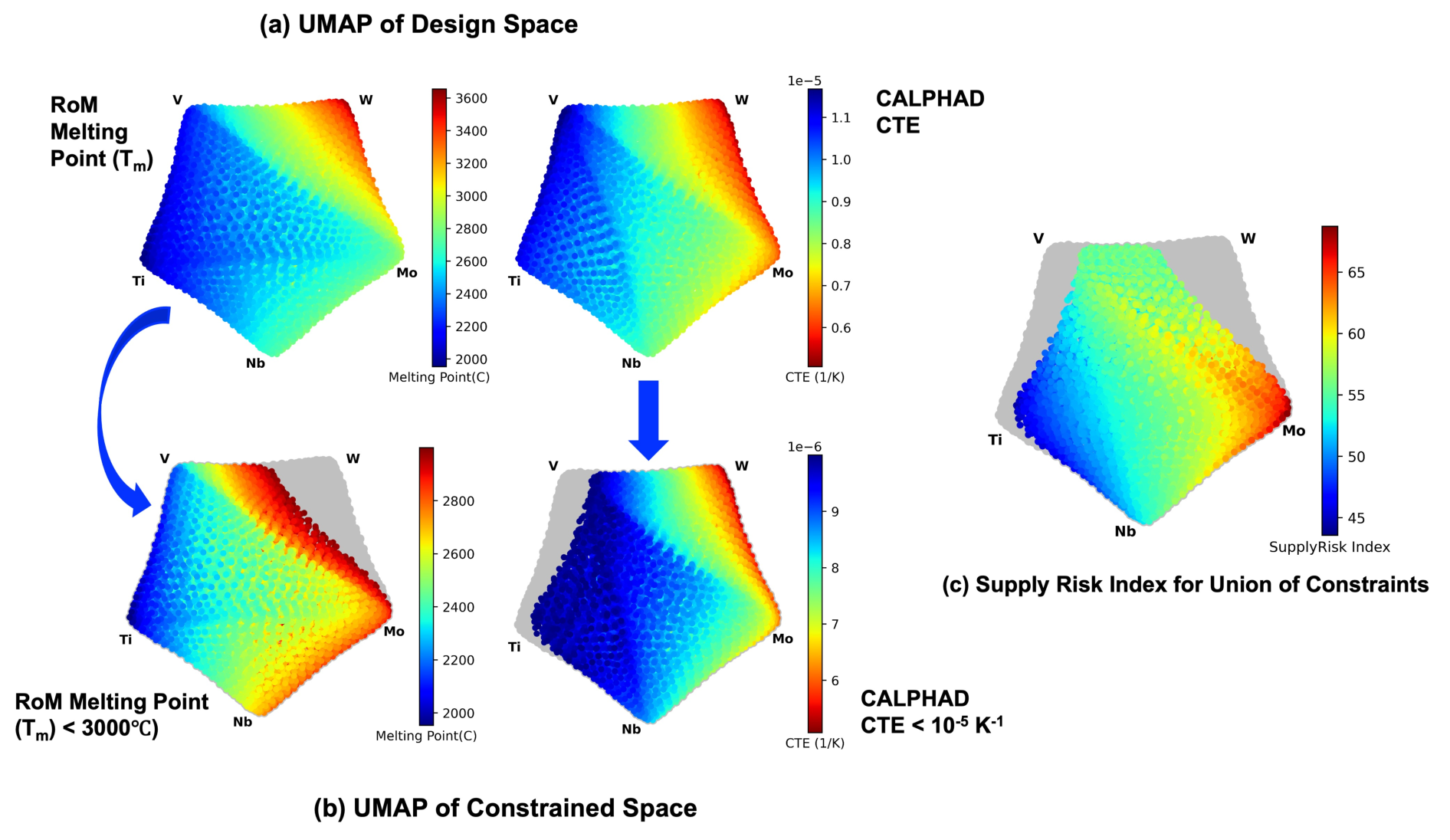}
        \caption{(a) UMAP of MoNbTiVW alloy space that depicts Rule of Mixtures Melting Point, CALPHAD Coefficient of Thermal Expansion (CTE). (b) UMAP\cite{VelaVisualizingPractices,Vela2024VisualizingPractices,vela2024visualizingarxiv} of MoNbTiVW alloy space that depicts Melting Point constraint, CTE constraint. (c)  Supply Risk Index for the feasible space with union of constraints}
        \label{fig:Constraints_UMAP}
    \end{figure}

For this study, the MoNbTiVW system and all its potential subsystems were sampled at 5 atomic \%, resulting in a design space of 10,621 candidate alloys. This design space was then constrained to satisfy the criteria described in Section~\ref{Methods} and \autoref{tab:constraint_objective}, reducing the feasible space to 7,991 candidate alloys. The constraints were evaluated using relatively inexpensive CALPHAD models and rule of mixtures approaches. A visualization of the design constraints applied in this study is shown in \autoref{fig:Constraints_UMAP}.

Constraining the design space based on melting point aligned with expectations, primarily eliminating compositions rich in tungsten (\autoref{fig:Constraints_UMAP}b). Similarly, constraining the design space based on the thermal expansion coefficient effectively excluded compositions rich in vanadium and titanium (\autoref{fig:Constraints_UMAP}b). Vanadium and tungsten have higher thermal expansion coefficients compared to the other elements in the system, and the elimination of these alloys is beneficial for effective thermal management and maintaining thermal stability. All alloys in the design space meet the full body-centered cubic (BCC) phase stability criteria.

To assess the impact and effectiveness of incorporating supply risk considerations into the design process, we conducted four distinct case studies, as detailed in the Methods section. The specific scenarios examined include:

\begin{itemize}
    \item \textbf{Design Space: Performance} - This case study focuses on optimizing application-based performance within the entire design space of 10,621 alloys, aiming to identify the best-performing solutions.
    \item \textbf{Design Space: Performance + Supply Risk Criteria} - Both performance metrics and supply risk indices are considered across the entire design space to balance high performance with low supply risk, integrating supply chain considerations into the design process.
    \item \textbf{Feasible Space: Performance} - This study narrows the focus to a subset of the design space that meets predefined feasibility constraints, optimizing solely for performance within this feasible space.
    \item \textbf{Feasible Space: Performance + Supply Risk} - Similar to the second case study, but within the constrained feasible space, this scenario aims to achieve optimal performance while minimizing supply risk among 7,991 feasible alloys.
\end{itemize}

When the four different scenarios were subjected to MOBBO, starting with 10 candidate alloys per batch per iteration, the framework effectively optimized and converged in as few as 7 iterations. In each case study, this resulted in the exploration of 70 candidate alloys demonstrating the efficiency of the BBO framework in rapidly navigating the design space and identifying optimal solutions with a relatively small number of iterations. From the BBO perspective, the convergence criterion ensures that the optimization process continues until the improvement in hypervolume becomes marginal, indicating that the search is nearing the optimal Pareto front. The convergence of the BBO framework was determined by HVI achieved from the previous iteration. By the 6th iteration, hypervolume improvement is less than 5 \% \emph{indicating that the rate of improvement has gradually decreased}. The reduction in HVI suggests that the optimization process has effectively explored the design space. The identified solutions are likely to be close to the optimal trade-offs among the chosen objectives in each case. Interestingly, in the case studies incorporating supply risk as an objective, the framework converged in 6 iterations, compared to 7 iterations for the case studies without supply risk. 

By incorporating supply risk as a new objective function, some design regions are explored that may have been discarded by the MOBBO framework if evaluated solely on performance. This additional objective allows for the identification of solutions that balance performance with supply risk, revealing new regions where reduced supply risk can be achieved, albeit at the expense of some performance degradation.

However, introducing an additional objective---especially one that often conflicts with others, like supply risk---adds computational complexity to the Bayesian-based optimization scheme. For instance, alloys with significant molybdenum (Mo) content tend to have higher costs and supply risks. Mo, a key component in the MoNbTiVW system, contributes to excellent high-temperature performance and mechanical strength, but its high cost and associated supply risk impact overall material sustainability. This trade-off between performance and supply risk requires careful consideration in the alloy design process, as demonstrated in our results. Incorporating supply risk as an objective ensures that some Pareto-optimal designs are not only high-performing but also sustainable and resilient from a supply chain perspective.

\begin{figure}[H] 
        \centering
        \includegraphics[width=0.9\textwidth]{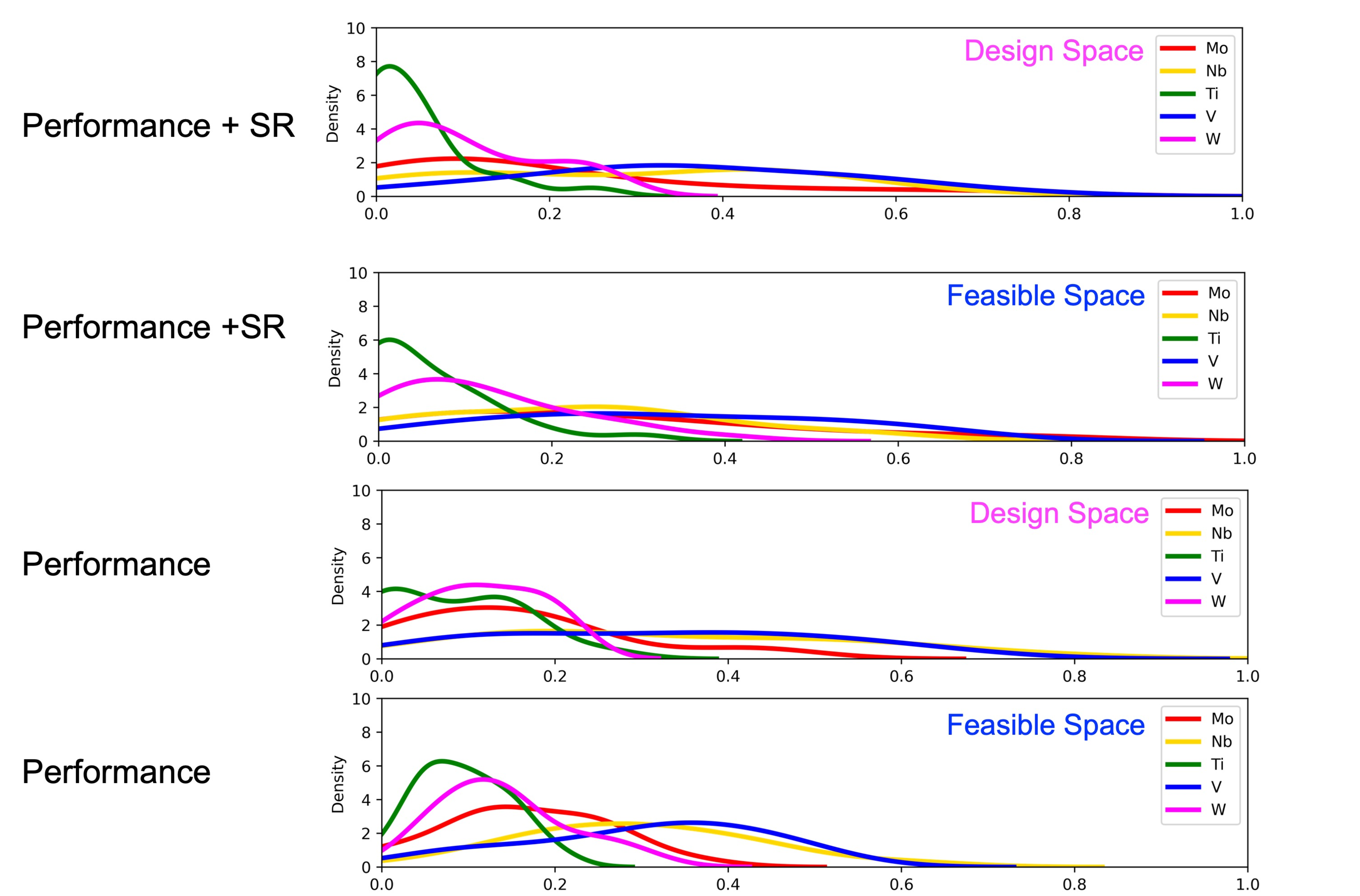}
        \caption{Chemical Signature Plots of Pareto Optimal Alloys showing exploration in each case study}
        \label{fig:Chemical_Signatures}
    \end{figure}

The chemical signature plots for all four case studies provide valuable insights into the exploration of the design space under the four scenarios considered (\autoref{fig:Chemical_Signatures}). In the optimization of case studies for both performance and supply risk considerations, the exploration predominantly centered on V, Nb, and Mo-rich alloys. This differs from the performance-only studies that mainly explored Nb and V-rich alloys. The efficient exploration across all scenarios indicates the effectiveness of the optimization approach in navigating the design space. It has been noted that Mo-rich alloys in the studies aimed at performance and supply risk scenarios suggest that the algorithm is attempting to learn and understand the boundaries of the design space. This observation highlights the algorithm's adaptive nature, potentially driven by the need to balance supply risk while optimizing performance metrics. This behavior underscores the importance of including supply risk in the optimization framework to uncover alloy compositions that may or may not have the best performance metrics but are more sustainable options.

\begin{figure}[H] 
        \centering
        \includegraphics[width=0.9\textwidth]{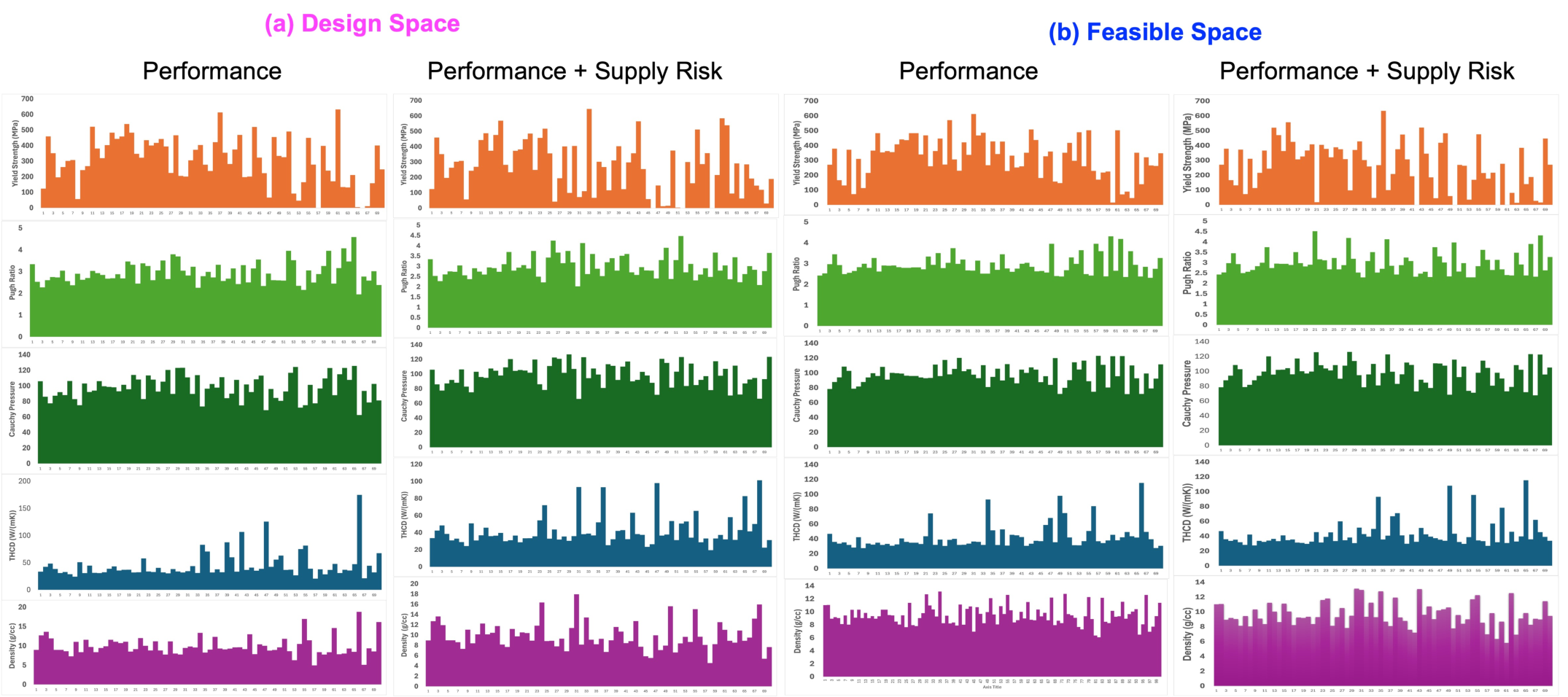}
        \caption{Summary of data for performance objectives for (a)Design Space (b)Feasible Space, showing contribution of each objective to hypervolume improvement}
        \label{fig:Objectives_Summary}
    \end{figure}

Based on the summary data for performance objectives optimized across all case studies (\autoref{fig:Objectives_Summary}), we observe that constraint satisfaction has proven to be an efficient approach, leading to better candidate suggestions even in the early iterations. The summary data also indicates that density has contributed less to the improvement in hypervolume compared to other objectives. Yield strength, ductility, and thermal conductivity often involve trade-offs. For instance, increasing yield strength can reduce ductility, while optimizing thermal conductivity can impact both yield strength and ductility. These trade-offs are crucial in determining the overall performance of the alloy, making them key factors in the optimization process. In this alloy system, higher density might correlate with certain desirable properties, such as high-temperature strength, but it must be balanced with considerations for ductility and thermal conductivity to achieve an optimal alloy design.

\begin{figure}[H] 
        \centering
        \includegraphics[width=1.0\textwidth]{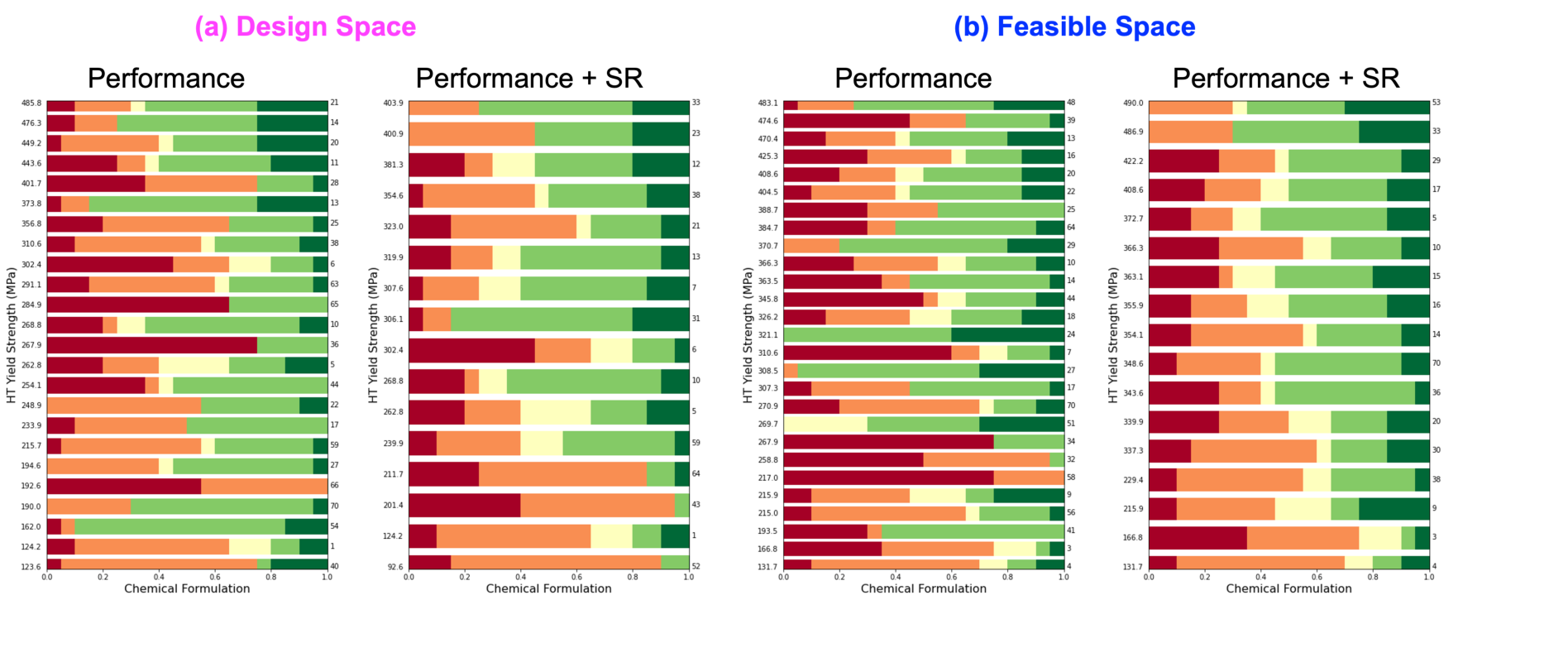}
        \caption{Compositional bar chart for Yield Strength of pareto optimal points for (a)Design Space (b)Feasible Space}
        \label{fig:YS_Summary}
    \end{figure}

Summary data clearly demonstrates that incorporating supply risk considerations into the optimization process maintained performance levels. To illustrate this, we will use yield strength of Pareto optimal alloys in all four case studies (\autoref{fig:YS_Summary}). The compositional bar chart of yield strength, which displays the distribution of yield strength among the Pareto optimal alloys, across the four different scenarios---design space focusing on performance alone, design space including both performance and supply risk, feasible space focusing on performance alone, and feasible space incorporating both performance and supply risk---the results show good consistency. These charts clearly show that the average yield strength is very similar across all scenarios. 

In contrast, for case studies that focused on performance alone, a significant number of Pareto optimal alloys are Mo-rich, which is associated with a higher supply risk index. Without considering supply risk, the optimization tended to favor compositions with higher Mo content due to its beneficial impact on performance metrics like yield strength and high-temperature stability. However, this preference for Mo-rich alloys also results in higher overall costs, underscoring the importance of incorporating supply risk considerations to achieve a more balanced and cost-effective alloy design.

\begin{figure}[H] 
        \centering
        \includegraphics[width=1.0\textwidth]{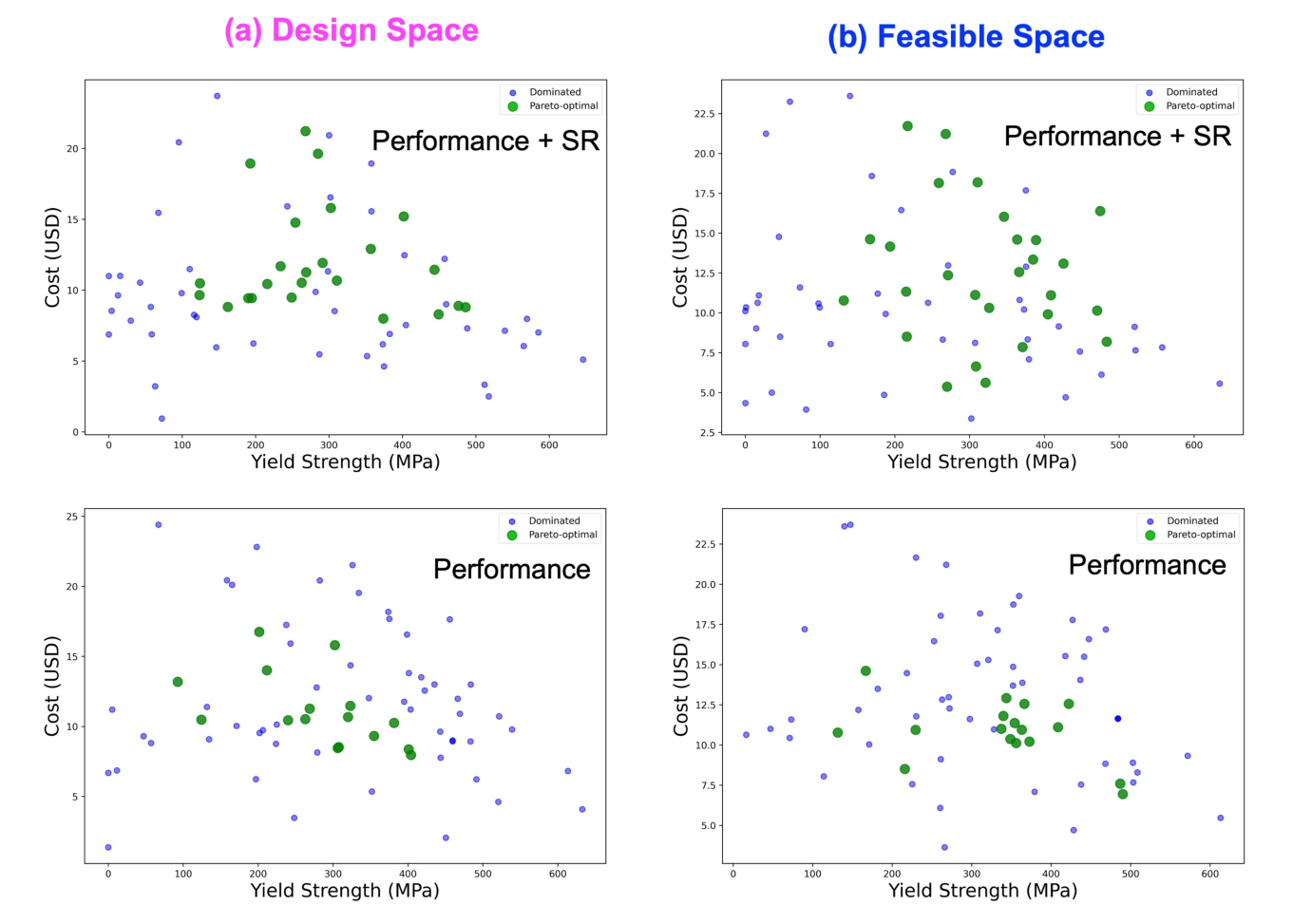}
        \caption{Pareto Plot for Yield Strength vs Cost (in USD) based on pure elemental raw material price of (a)Design Space (b)Feasible Space}
        \label{fig:YS_vs_Cost}
    \end{figure}

In addition to yield strength, the 2D plots of YS versus cost (in USD, per lb, based on the pure elemental raw material price) show that incorporating supply risk considerations allowed the algorithm to explore more low-cost options compared to solely focusing on performance (\autoref{fig:YS_vs_Cost}). While cost tends to be a lagging indicator for more complex supply chain risk scenarios, it is an easily understood factor that can be used to compare new designs to existing solutions.

By including supply risk as an objective, MOBBO framework explored different areas of the design space, including lower supply risk options, to identify Pareto optimal solutions that balance multiple objectives efficiently. This comprehensive exploration ensured that the final alloy designs also included solutions that meet desired performance metrics while being sustainable.

\section{Conclusion}

In summary, the results show that integrating supply risk considerations into the alloy design process \textbf{does not result} in a trade-off with performance metrics. Exploration of different scenarios proved that multi-objective batch Bayesian optimization (MOBBO) framework is very capable of managing multiple, potentially conflicting objectives simultaneously. By incorporating supply risk, the framework successfully balanced this additional objective with performance metrics like yield strength, \emph{finding solutions that did not compromise critical performance attributes}. Moreover, 2D plots of YS versus cost indicate that supply risk integration allowed the algorithm to explore more low-cost options compared to focusing solely on performance. Interestingly, case studies focusing only on performance tended to favor Mo-rich (higher cost) alloys, emphasizing the need for supply risk considerations to achieve a more balanced and cost-effective design. Second, the efficient management of constraints related to supply risk—such as melting point and thermal expansion coefficient—allowed the optimization process to focus on viable candidates without significantly affecting performance. Lastly, the integration of supply risk into the design process enabled the identification of alloys that not only met performance requirements but also exhibited lower supply risk. The minimal variation in average yield strength across the case studies indicates that the optimization process effectively identified high-performing alloys while incorporating supply chain concerns. Additionally, the ongoing development of large language models (LLMs) for automated indicator extraction will be very useful in identification and assessment of relevant sustainability and supply chain indicators. 
The consistent average YS across scenarios, the effective management of constraints, and the comprehensive exploration of design space underscore the robustness of the MOBBO framework in addressing complex multi-objective optimization problems in alloy design. Integrating supply risk considerations as part of design and development makes optimization process more holistic and necessary in real-world applications.

\subsection*{Future Work and Challenges}
While the current study has successfully demonstrated the benefits of incorporating supply risk into the alloy design process, several challenges and areas for future work remain. One major challenge is the computational complexity introduced by additional objectives, such as supply risk, which necessitates more sophisticated and efficient optimization algorithms. Future work should focus on developing advanced algorithms that can handle multiple conflicting objectives more efficiently and reduce computational overhead. Another area for future research is the expansion of the optimization framework to include other sustainability metrics, such as environmental impact and recyclability. Integrating these additional objectives will provide a more comprehensive assessment of the sustainability of alloy designs. Another important addition would be the integration of automated indicator extraction tool into the optimization framework to enhance its capability to handle complex data and extract meaningful insights, further improving the alloy design process.

A significant advantage of this approach is the incorporation of a supply risk-aware design strategy as an inherent part of the development process. This ensures that supply chain considerations are systematically integrated from the outset, leading to more sustainable and economically viable alloy solutions. In conclusion, while this study demonstrates the ability to integrate supply risk considerations into alloy design, ongoing advancements in computational methods, optimization algorithms, and automated data extraction tools will be crucial in addressing the remaining challenges and further improving the sustainability and performance of future alloy design.

\section{Acknowledgements}

Research was sponsored by the Army Research Laboratory and was accomplished under Cooperative Agreement Number W911NF-22-2-0106 as part of the BIRDSHOT Center within the High-throughput Materials Discovery for Extreme Conditions (HTMDEC) program. We  acknowledge the support from NSF through Grant No. NSF-DMREF-2323611 (\emph{DMREF: Optimizing Problem formulation for prinTable refractory alloys via Integrated MAterials and processing co-design (OPTIMA)}) and grant number 1746932. We acknowledge grant number NSF-DGE-1545403 (\emph{NSF-NRT: Data-Enabled Discovery and Design of Energy Materials, D$^3$EM}). The authors would also like to acknowledge TAMU High Performance Research Computing group for access to computing resources needed for the calculations in this project. We also acknowledge the use of OpenAI's ChatGPT-4 for large language model development and some sentence structure formations in the manuscript. After using this tool/service, the author(s) reviewed and edited the content as needed and take(s) full responsibility for the content of the publication.  

\section{Data availability}

Code and data generated in this work will be available upon request.

\section{Compliance with Ethical Standards}

Conflict of Interest: On behalf of all authors, the corresponding author states that there is no conflict of interest.

\section{Supplemental Information}

Code for this project is available on GitHub at https://github.com/mrinalinimulukutla/SupplyRiskAwareDesign and is available upon request. 

 \subsection{Appendix-1}
 Supply Risk Indicator Data for different elements has been collected and tabulated at \href{https://docs.google.com/spreadsheets/d/170n_NIb86QLPdVgY1ByG-osbH0xV2FY1/edit?usp=drive_link&ouid=105603584489967562440&rtpof=true&sd=true}{the link attached}, Table S1: Elemental Info, Table S2: Calculation Detail, Table S3: HHIL Calculation, Table S4: Element Price History

 \subsection{Appendix-2}
 \emph{ Large Language Models for Indicator Extraction}

The acquisition of data required to compute the supply risk of a given element is a time-consuming process. The statistics that inform supply risk indicators are derived from documents, periodicals, and research papers, which vary in length, language, and format. Traditionally, accurate data collection necessitates manual searching and sorting by a human to extract relevant information and compile it into a usable table for research purposes. However, recent advancements in Large Language Models (LLMs) and Natural Language Processing (NLP) have enabled the automation of this task. 

A workflow leveraging LLMs has been defined, and an application is currently under development to support the automated extraction of statistics for supply risk indicators. The objective of this workflow is to minimize human involvement in data collection for supply risk indicators while ensuring accuracy and allowing for easy modifications to the list of indicators used in calculating the supply risk for a material. The method for data extraction is twofold: deterministic text search algorithms are employed to identify areas where relevant statistics may be found, followed by the use of a Large Language Model (LLM) for Natural Language Processing (NLP) to perform binary classification to locate important data.    
    
There are four main steps that define the workflow for data extraction. 

\emph{Document Retrieval and Processing}: The process begins with retrieving documents for a given indicator. Document locations can be specified through either direct links associated with each statistic or via search terms and a search URL. Direct links are particularly useful when the data source is already known or a specific source is preferred. Special handling is required for periodicals or cases where the publication year is part of the document link. In these cases, the process starts with the current year; if the document is unavailable, the year is decremented and the search is repeated until the document is found or the maximum iteration count is reached. This method ensures that the most recent publication is used for data extraction. The code for this function is provided below. It is important to note that this function is an opt-in feature---it triggers only when formatted correctly, and it avoids downloading duplicate documents.
    
    \begin{minted}[linenos, breaklines, bgcolor=white!20]{python}
    import requests
    from datetime import datetime
    from urllib.parse import urlparse
    import logging
    import os

    def download_periodical(start_date, url, max_iterations):
        logging.basicConfig(level=logging.INFO)
        year = start_date.year
        for i in range(max_iterations):
            current_year = year - i
            current_url = url.replace("@year", str(current_year))
            parsed_url = urlparse(current_url)
            file_name = parsed_url.path.split('/')[-1]
            file_path = os.path.join('sources', file_name)
            if os.path.exists(file_path):
                return file_path
            try:
                response = requests.get(current_url)
                response.raise_for_status()
            except requests.RequestException as e:
                logging.error(f"Request failed for year {current_year}: {e}")
                continue
            if response.status_code == 200:
                with open(file_path, 'wb') as file:
                    file.write(response.content)
                return file_path
            else:
                logging.info(f"Publication not found for year {current_year}, trying the previous year...")
        return False
        \end{minted}

An example where this approach is particularly effective is the United States Geological Survey's yearly Material Commodity Survey report. The link to the report can be constructed as:

\begin{center} \texttt{https://pubs.usgs.gov/periodicals/mcs\{year\}/mcs\{year\}.pdf} \end{center}

By iterating backward from the current year, the most recent available document is downloaded and used for data extraction.

The use of search terms and a search URL is currently under development and testing. This functionality aims to find data sources that are not published in periodicals but frequently provide relevant statistics, such as recently published research papers. The process involves searching a repository using keywords from the indicator definition, followed by text analysis techniques to identify results with a high likelihood of containing relevant information. LLMs may also be employed to classify the most pertinent results. Further details on this functionality will be included in a future publication.

\emph{Data Reduction}: After downloading the document, it is stripped down to text only, and various text search methods are applied to minimize the amount of text sent to the LLM. This step is crucial for reducing the number of LLM queries, which is the primary performance bottleneck in this workflow. Since querying LLMs can be costly depending on the model used, it is essential to limit reliance on LLMs while maximizing their effectiveness.
    
The first step is to collect keywords for the text-based search within the document. These keywords are used to identify the location of relevant data. Initially, the keywords for any given indicator will include the name of the material for which statistics are being extracted and the specific terms defined in the indicator's statistic. A key consideration is that keyword matching may fail if the document is written in a different language. To address this, the LLM is queried to detect the document's language and translate the keywords accordingly. The template for this query is provided below:

    \begin{center}
        \textit{Please translate these keywords into the language of the text included below in "". Keywords: \{keywords\}, Text: "\{text\}"}
    \end{center}

In this example, the text provided to the LLM consists of only the first ten sentences.

Next, a text search method is employed to identify sections of the document that may contain relevant data. The primary text search algorithm used is a keyword proximity search, which scans the document for sets of keywords that appear within a certain distance of each other. This general-purpose method is applicable to any indicator, with keywords defined to ensure that closely appearing words are likely related to the statistic being extracted. These identified sections are then added to a queue for LLM classification. Vector embeddings are another option for keyword search, and an implementation is currently being developed.

Another text search algorithm used is a header-searching algorithm, designed specifically to find headers that match the keywords. Utilizing PDF analysis tools, this method searches for isolated text near the top of the page that aligns with the keywords. This is particularly useful for documents where statistics are presented in tabulated form, such as the USGS MCS reports. All text on pages with matching headers is queued for LLM classification.

\emph{Large Language Model Processing}: All of the instances of possibly relevant text are passed to a LLM with the intent of implementing a binary classification. This acts as another filtering step to exclude as much irrelevant data as possible. To this end, the LLM is queried for each possible relevant text with the following template.
    \begin{center}
        \textit{Return True if this text reports a statistic for \{Statistic-name\}. Otherwise return False Text: \{text\}}
    \end{center}
The \textit{Statistic-name} is defined for each statistic within an indicator, and the text refers to the potentially relevant content from which data might be extracted. When the LLM returns \textit{true}, the text is retained; otherwise, it is discarded. The remaining text is then queried using the LLM, following the specific queries defined for each statistic in the indicators. Both the query and the expected return format are provided to the LLM, guiding it on how to format the extracted data.

\emph{Data Validation and Error Handling}: Validation begins by reviewing the outputs from the LLM after text extraction. There are several reasons why validation is necessary. First, multiple statistics may be identified through the text search and LLM classification. Since only one statistic can be used for calculating the indicator, any extraneous statistics must be filtered out. This is made possible by the LLM’s ability to make contextual decisions. By providing the LLM with the text from which the statistics were drawn, along with additional context, unnecessary statistics can be eliminated.

Second, there may be instances where the text search fails to locate relevant information. In such cases, the document should be manually reviewed to ensure no data is missed. If the workflow fails to identify relevant data, adjustments to the text search methods may be needed, as discussed in the challenges section.

Third, there may be situations where the LLM is unable to extract information from the relevant data sources. This issue often arises from prompt engineering, which the user can address by editing the prompts included in the indicator definition. By accounting for these potential errors, the following cases are defined and managed accordingly.

\begin{itemize}
    \item \emph{Case A}: Only one statistic is found, and it is successfully returned. This is the ideal scenario, where the statistic is reported only once, and the LLM extracts it without issue. The statistic is then passed to the calculation logic.
    
    \item \emph{Case B}: Multiple statistics are found and successfully extracted. In this situation, the statistics are compared. If they are identical across all instances, the statistic is recorded, and the process continues. If there are discrepancies, the text from each query is combined, and the LLM is queried to determine which statistic is the most recent and accurate one, with the current date provided for context. The results of this query are validated again, and only if a single statistic is identified and extracted is it recorded. Otherwise, the failure is reported, and the indicator is skipped in the supply risk calculation.
    
    \item \emph{Case C}: No statistics are found, or no values are successfully returned. In this case, the failure is logged, the user is notified, and the indicator is skipped in the overall supply risk calculation.
\end{itemize}

\emph{Indicator Calculation}: The value for each statistic is recorded, and the process is repeated for all other statistics listed in the indicator. Once all statistics are recorded, the indicator is calculated using the formula provided in the indicator definition. This process is repeated for each defined indicator, and the overall supply risk indicator for the material is then calculated. The final result, along with the statistics used in the calculation and logs detailing all the steps, is returned to the user.

\emph{Indicator Definition}: To support the proposed workflow, indicators are defined with fields for name, statistics, and calculation. These indicators serve as the primary tool by which the statistic extraction is conducted. The intended use case is for the user to edit to, add, or delete these indicators to control the calculation of the supply risk for a material. Furthermore, the workflow should not require further interaction from the user besides the definition of the indicators. To accomplish this, the indicators are defined very flexibly to support many cases, and the workflow is modelled such that it can handle many cases for types of documents that statistics may be drawn from. The indicator name is just a label used for reporting errors to the user. The statistics field holds a list of statistic objects which the code will iterate through. Each statistic object holds the statistic name, which serves as a reference for the calculation, the data source, and a prompt for binary classification while searching text. There is a prompt for text extraction, which is paired with a JSON format that details how the extracted data should be returned to the application. Multiple statistics can be specified to satisfy the requirements for the calculation of the indicator. The final section of the indicator definition is the calculation. This is a mathematical formula using the names of the statistics defined previously.

A key aspect of this workflow is ensuring that indicators are general-purpose. Ideally, an indicator definition should remain functional over time without requiring frequent updates, and it should work across various elements or materials without the need for manual edits in each case. To achieve this, the indicator definition supports the use of placeholders labeled \texttt{"@Element"} and \texttt{"@Year"}. These placeholders allow the application to automatically insert the relevant data into prompts and URLs before using them. For documents published in annual reports, this feature enables the URL to be adjusted automatically to retrieve the most recent publication available.

Similarly, prompts sent to the LLM utilize the element or material name to locate relevant statistics. This approach removes the need for users to create specific prompts for each material tailored to the data source. However, it may introduce some inconsistencies when data for different materials in the same document are reported in varying formats.

Constructing these indicators requires substantial development effort, and careful prompt fine-tuning is necessary to achieve consistent results from the LLM-enabled text extraction. However, the outcome is a near-autonomous extraction of relevant statistics for each indicator. This approach enables the inclusion of any number of indicators in the overall calculation process without the need to modify the underlying code. Essentially, to add a new indicator, the user only needs to create a new JSON file to be included in the application.

\bibliographystyle{elsarticle-num}
\bibliography{mybibfile}

\end{document}